# Prediction of the energy and exergy performance of F135 PW100 turbofan engine via deep learning

Mohammadreza Sabzehali[a], Amir Hossein Rabiee[b], Mahdi Alibeigi[a], Amir Mosavi[c]

[a] Department of Mechanical Engineering, Tarbiat Modares University, PO Box 14115-143, Tehran, Iran
[b] School of Mechanical Engineering, Arak University of Technology, 38181-41167 Arak, Iran
[c] Faculty of Civil Engineering, TU-Dresden, Dresden, Germany

**Abstract**

In present study, the effects of flight mach number, flight altitude, fuel types, and intake air temperature on thrust specific fuel consumption (TSFC), thrust, intake air mass flow rate, thermal and propulsive efficiency, as well as the exergetic efficiency and the exergy destruction rate in F135 PW100 engine are investigated. Based on the results obtained in the first phase, to model the thermodynamic performance of the aforementioned engine cycle, flight mach number and flight altitude are considered to be 2.5 and 30,000 m, respectively, due to the operational advantage of flying at ultrasonic altitude, and higher trust of hydrogen fuel. Accordingly, in the second phase, taking into account the mentioned flight conditions, an intelligent model has been obtained to predict output parameters (i.e. thrust, thrust specific fuel consumption (TSFC), and Overall exergetic efficiency) using the deep learning method. In the attained deep neural model, the HPC pressure ratio, fan pressure ratio, turbine Inlet temperature, intake air temperature, and bypass ratio are considered as input parameters. The provided datasets are randomly divided into two separate sets: the first set contains 6079 samples for model training and the second set contains 1520 samples for testing. Particularity, the Adam optimization algorithm, the cost function of the MSE, and the active function of Relu are used to train the network. The results show that the error percentage of the deep neural model is equal to 5.02%, 1.43%, and 2.92% in order to predict thrust, TSFC, and Overall exergetic efficiency, respectively, which indicates the success of the attained model in estimating the output parameters of the present problem.

**Keywords:** Dual spool turbofan; F135 PW100; Mixed-flow turbofan; Deep learning; Energy and exergy.

# 1. Introduction

Gas turbines (GT) are one of the powers generation cycle types that is an internal combustion engine (ICE) of a rotary machine. These engines operate on the Brayton cycle (BC). Classically, the GTs have extensive applications in various industries ranging from oil, gas, and petrochemicals to power generation plants and various propulsion systems like airplane propulsion structures.

The simplest GT engine configuration is Turbojet (TJ). In the TJs, the air is first entered into the compressor, then it enters the combustion chamber, after that, it increases its temperature and pressure, subsequently, it enters the turbine and it decreases its temperature and pressure. The power turbine supplies the mechanical power required by the compressor. The emission gases generated from the combustion of fuel and air after exiting the turbine are reflected in the nozzle. So, the flow velocity increases, then it depletes to the ambient. Due to the discharge of the turbine output flow, it is generated proportional to the velocity and pressure of the flow in the nozzle output and the ambient pressure.

One of the most essential aero gas turbine engine types is the turbofan engine. The internal flow is separated into two fragments: the hot flow and the bypass flow in all the turbofan engines. Turbofan engines are divided into two types: twin-spool, and three-spool. Correspondingly, Turbofan engines syndicate bypass flow as cold flow, and turbojet as the hot core [1, 2]. Another division of turbofan engine includes two types: the mixed-flow turbofan engine and unmixed-flow turbofan jet engine. In mixed turbofan, the mixing of the intake airflow into the hot core enters the combustion chamber. The fuel combustion products go into a high-pressure turbine (HPC) and then push through the low-pressure turbine (LPC) after crossing HPC. So, the output flow enters from the low-pressure turbine (LPC) to the mixer and it combines with the bypass flow or fan output flow. Subsequently, the mixer output flow arrives at the nozzle. Finally, it exhausts to the ambient.

Recently, many studies have been done to improve turbofan engine performance. For example, Balli et al. [3] examined performance parameters, environmental sensitivity, and sustainability of TF33 turbofan engine exploited widely in military aviation. In their investigation, Valuation parameters including energy efficiency, exergy destruction ratio, fuel heating value ratio (FHVR), specific fuel consumption (SFC), and Thrust, and so on were considered. Chen et al. [4] numerically considered a turbofan with an inlet ejector nozzle fortified with a supplementary inlet door. This supplement has been used in the inlet bypass flow to ejector the nozzle. It increased the engine installed Thrust. Also, the ejector nozzle model is applied to analyze the performance of its effect on the engine. Moreover, a prediction method has been proposed for the exhaust system of backward infrared radiation intensity prediction. The results indicated that the turbofan engine with an inlet ejector not only decreases the infrared radiation and specific fuel consumption (SPFC) compared to the conventional turbofan engine but also improved significantly the engine installed Thrust, so the engine performance has been developed. Xu et al. [5] evaluated a new mixed-flow turbofan and called a novel re-cooled mixed-flow turbofan cycle (RMTC). It has been used for military aero-engine with a high Thrust-weight ratio. Their investigation indicated that the specific Thrust has been improved by increasing the outer fan flow temperature. Similarly, the turbine cooling air consumption can be concentrated by reducing the inner fan flow temperature during the compression process just by adding a re-cooler. Also, they investigated the effects of bypass ratio on the specific Thrust, fuel consumption rate, and re-cooler location on the RMTC performances in the whole applicable fight conditions. The parametric studies approved that the specific Thrust enhancement of RMTC is more substantial at high flight-Mach number. Finally, the calculation results confirmed that the considered RMTC system proves potential application for high-speed military aircraft. Rao et al. [6] studied to improve the propulsive efficiency of a civil aero-engine design aimed at lowering specific Thrust and idle descent conditions by swelling the bypass ratio. Also, the bypass ratio of discharge and core nozzles of a high-bypass ratio aero-engine have been studied in isolation and installed on an airframe. The consequence was showed that the supreme alteration in the bypass nozzle discharge coefficient between the installed and isolated aero-engine across the descent phase is $\simeq 1.6\%$. Bali and Caliskan [7] premeditated the JT15D turbofan engine and all of its components. They intended on energy, exergy, environmental, aviation, and sustainability analysis. Firstly, They calculated the system's specific Thrust, the specific fuel consumption of the engine, the system's energetic efficiency, the exergetic efficiency, the system's exergetic improvement potential rate, productivity lack ratio, and fuel exergy waste ratio with amounts of 315.9 N.s/kg, 15.8 g/kN.s, 21.15%, 19.919%. as 1573.535 kW, 402.024%, and 80.081%, respectively. Balli et al. [8] thermodynamically analyzed the TF33 turbofan engine fueled by hydrogen with conflicting to kerosene. The results showed that the fuel flow, the specific fuel consumption, energy efficiency, and thermal limit ratio were decreased by 63.83%, 60.61%, 0.757%, and 1.55%, respectively. In another study, Akdeniz and Balli [9] analyzed energy, exergy, and sustainability analysis for the PW4056 turbofan engine and its main components to detect the different fuel impacts at the same dead state conditions. They understood that the amount of fuel mass flow and the exhaust gases mass flow of the hydrogen fuel is lower than the kerosene fuel with the value of 1.03 kg/s, and 2.85 kg/s for the hydrogen fuel and 117.14 kg/s, and 118.96 kg/s for the kerosene fuel, respectively. In both cases, the minimum exergy efficient component calculated 64.24 % on the combustion chamber (CC) for the kerosene case and 58.20 % for the hydrogen case. The lowest relative exergy loss ratio was determined to be 28.28 % for Fan outlet loss for hydrogen case, even though the maximum relative exergy consumption ratio was determined to be 51.93 % for CC components for hydrogen case.

Thermodynamic analysis of gas turbines is done to evaluate the performance of these engines by design variables. In this regard, Ibrahim et al. [10] proposed energetic analysis and exergitic analysis of a gas turbine plant cycle. They evaluated the effect of exergy flow, inlet flow, and outlet flow of compressor, turbine, and combustion chamber in terms of physical exergy and chemical exergy. Their results showed that the highest exergy destruction rate was associated with the combustion chamber. In another investigation, Zhao et al. [11] analyzed the first and second laws of thermodynamics analysis for Intercooled Turbofan Engine (ITE) in different working conditions. Their consequences showed that the highest exergy destruction between engine components is for combustion chamber also, the intercooler method is a cause of Exergy destruction decreasing in the combustion chamber. Aygun and Turan [12] studied the Exergy performance of the gas turbine variable cycles for the next-generation combat aircraft. Also, considering the bypass ratio and the input temperature of the turbine as design variables, these variables were optimized using the genetic algorithm to maximize fuel consumption. They calculated the lowest possible amount of fuel consumption as 17.41 grams per second.

Najjar and Balawneh [13] thermodynamically analyzed and optimized turbojet propulsion as one of the gas turbine (GT) types. In their study, the special Thrust is strongly dependent on turbine inlet temperature (TIT). So, a 10 percent reduction in TIT leads to a 6.7% reduction of the specific Thrust and a 6.8% reduction in Specific Fuel Consumption (TSFC). It is also optimal values for the Turbojet parameters in the flight altitude condition of 13,000 meters and the Mach number of 0.8 with the compression ratio of 14 and TIT of 1700 K. Also, Hendricks and Gray [14] produced a new tool for analyzing and optimizing the thermodynamics performance of GT-cycles. They called the device as pyCycle. The code can compute the data of thermodynamic cycles at a rate of 0.03 % compared to the results obtained by the NPSS program. In another study, Xue et al. [15] studied the effects of bypass ratio in the optimal compression ratio of fan in the Trent - XBW engine series have been investigated. It showed that the bypass factor is a cause of more effective in reducing fuel consumption and increasing the Thrust force.

Previous studies show that molecular weight and the thermal value of fuel affect the performance of gas turbine engines. Also, Exergy analysis, the thermal value of the engine, and the chemical formula of fuel are effective on the Exegetic operation of gas turbine engines. In recent studies, the effect of the use of Hydrogen fuel compared to hydrocarbon fuels on the performance of gas turbine engines is investigated. Balli et al. [16] investigated the effect of using hydrogen fuel on the Exergetic performance of a turbojet engine. Their results showed that the using of hydrogen fuel compared to hydrocarbon fuels, the exergy efficiency was reduced from 15.40% to 14.33 %. In another study, Gaspar and Souca [17] considered the effects of different fuels on the functional performance and environmental performance of a small turbofan engine. Also, the performance and intensity of ecological performance were investigated in the different working conditions. Verstraete [18] compared the action of the fueled hydrocarbon and fueled Crocin of long-range aircraft. By using hydrogen as fuel compared to the utilization of Crocin fuel, the direct operating cost and energy consumption have decreased by 3 and 11 %, respectively. Derakhshandeh et al. [19] simulated and optimized the environmental analysis and economic analysis of the GE90 turbofan engine. In their study, a comparison between hydrocarbon fuels and hydrogen fuel has been performed on design conditions. In Hydrogen fuel, the GE90 turbofan engine economically and environmentally has been optimized. Their consequences showed that the optimized cycle has been increased the Thrust force and thermal efficiency 16.27% and 2.65 %, respectively by using Hydrogen fuel; also, propulsive efficiency and overall efficiency decreased by 2% and 2.5%, respectively.

Recent studies indicate that the variation of the input temperature affects the performance of the energy of gas turbine engines. The variation of air temperature changes causes the variation of the incoming air density and changes in the inlet airflow to the engine. Changes in the inlet air inlet on the engine cause the switching performance parameters and exergy efficiency of gas turbine engines. Also, Caposciutti et al. [20] studied the effect of ambient temperature changes on the performance of a gas turbine plant with biogas fuel. In their study, by reducing the input temperature, the Power plant increased 4.5 %.

In addition to operational advantages due to the reduction in the dynamic pressure of the airflow and the bulk of the drag force and the Reynolds number, it is crucial to assess the challenges of the gas turbine engine cycle in high altitude flight conditions. In the study of Treuren and McClain [21], the values found in the fan pressure ratio and the bypass ratio for a turbofan engine are engines designed to fly at the height of 65,000 feet above sea level. The fan pressure, compressor pressure, and bypass ratio have been calculated at 1.57, 16.7, and 5.45, respectively.

In recent years, the importance of intelligent systems and embedded systems are the cause of intending the deep learning and neural intelligence system to control and intellect any systems. In this regard, many studies are worked on the aero system based on intelligent systems.

Kaba et al. [22] dramatically enhanced both civil and military aircraft engines with the assistance of an improved least-squares estimation-based genetic algorithm (LSEGA) in fight phases. The parametric studies such as Thrust, specific fuel consumption (SFC), overall and exergy efficiencies are considered. The results of the off-design parameters study verified that the root-mean-square (RMS) error was calculated as low as 0.000162. Also, RMS of exergy efficiency, overall efficiency, Thrust, and SFC determined 0.000311, 1.007, and 0.0763, respectively. Eventually, they observed that the considered LSEGA algorithm has effectively converged into optimal solutions for all indexes, and fight conditions with high accuracy.

Wang et al. [23] forecasted with remaining useful life (RUL) for the aircraft to reduce maintenance costs and develop maintenance strategies cause of expensive components. They proposed a novel concurrent semi-supervised model (NCSS) to estimate the RUL of the aero-engine. The NCSS could be provided satisfying prediction consequences with only a small amount of labeled data. The experimental fallouts indicated that the considered method was operative in the commission of RUL.

The above literature review shows that, while there exists a large body of literature studying the thermodynamic analysis of gas turbine engines, it appears that the comprehensive investigation has not been considered on the F135 PW100 engine. Accordingly, in the present study, for the first time, the effects of flight-Mach number, flight altitude, fuel types, and intake air temperature on thrust specific fuel consumption (TSFC), intake air mass flow rate, thermal and propulsive efficiency, the exergetic efficiency, and the exergy destruction rate in F135 PW100 engine are investigated. Moreover, based on the results obtained in the first part, a deep neural model for predicting the thrust, TSFC, and Overall exergetic efficiency is attained based on the deep learning approach.

## 2. Problem description and modeling

*2.1. Energy and exergy modeling*

A schematic of the dual spools mixed-flow turbofan engine configuration with inlet air cooling system illustrates in Figure 1. In this system, the input airflow is driven into the fan from the inlet air cooling system. A portion of the fan output flow enters the bypass channel and enters the mixer. The other part of the fan output flow enters into a high-pressure compressor (HPC). The HPC output flow enters the combustion chamber (CC), and the air reacts with fuel in the CC. Subsequently, the flow of air combustion products with fuel enters a high-pressure turbine and drives it to motion. The output flow of the high-pressure turbine enters the weaker pressure turbine named a low-pressure turbine (LPC) and then enters the mixer. Subsequently, the bypass channel output flow and the output flow of the LPC are combined in the mixer, then the mixer output flow is produced by the nozzle; finally, it exhausts to the ambient.

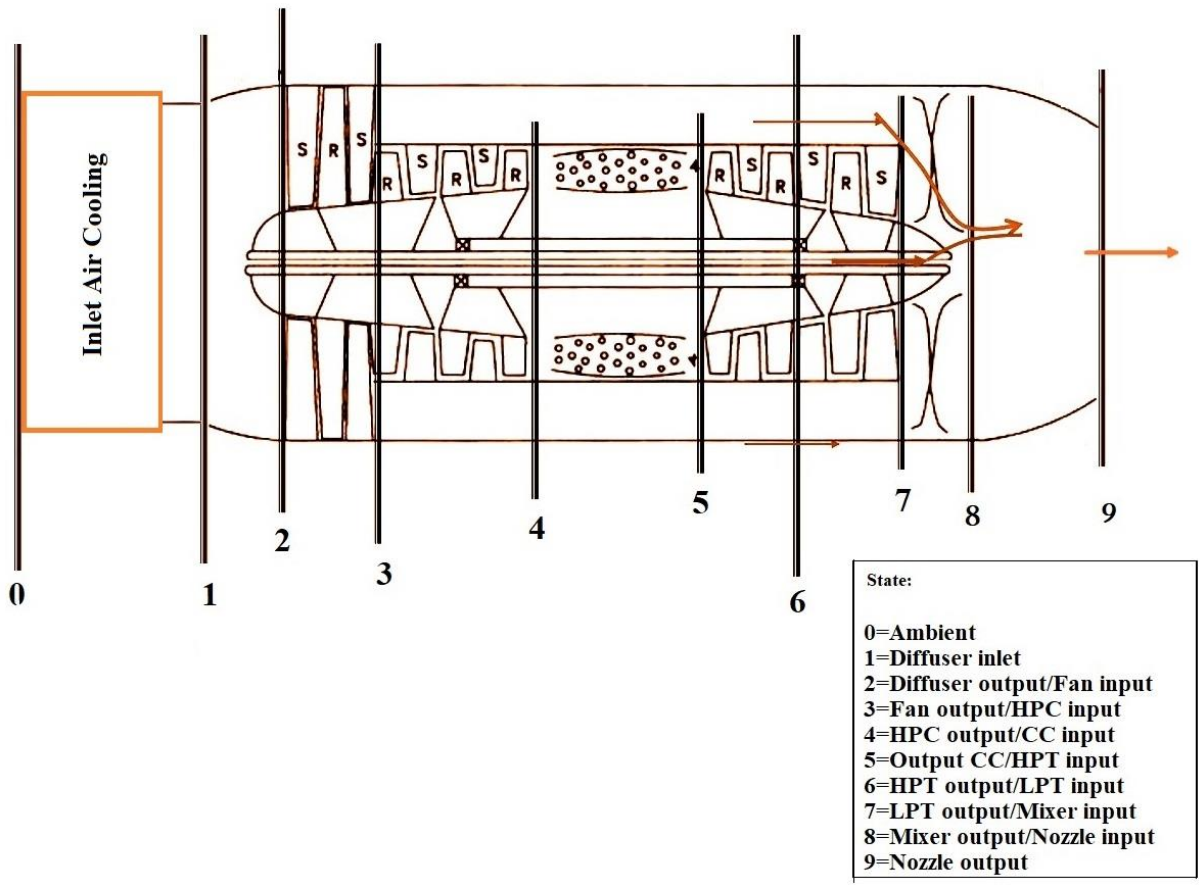

**Fig. 1.** schematic of the dual spools mixed-flow turbofan engine configuration with inlet air cooling system.

The Governing energy analysis of this study is presented as follows. The changes in airflow pressure at the intake and cooling system are negligible, so $P_0 = P_1$. It shows that the diffuser inlet pressure equals ambient air pressure. Also, the inlet temperature can be calculated as;

$$T_1 = T_0 + \Delta T \tag{1}$$

where T₀ and ΔT are ambient air temperature, and difference inlet temperature, respectively. Assuming the diffuser is isentropic, the diffuser exit air temperature is calculated as follows;

$$T_2 = T_1\left(1 + \frac{k_d - 1}{2}M_a^2\right) \tag{2}$$

where $T_1$ and $T_2$ are exit and inlet air temperature to the diffuser $k_d$ Specific heat at constant pressure to specific heat in constant volume for air passing through the diffuser, which is a function of the average temperature of the air passing through flight-Mach number. $M_a$ is the flight Mach number. The output air pressure of the diffuser is calculated as follows;

$$P_2 = P_1\left(\frac{T_2}{T_1}\right)^{\frac{k_d}{k_d-1}} \tag{3}$$

where $P_1$ and $P_2$ are air pressure at exit and inlet of the diffuser, respectively; and $T_1$ and $T_2$ are exit and inlet air temperature to the diffuser. $k_d$ specific heat at constant pressure to specific heat in constant volume for passing air of diffuser which is calculated as a function of the average temperature of the passing air of the diffuser. Assuming air and fuel combustion products with air are ideal gas. The fan output pressure is obtained from the following equation.

$$P_{3=\pi_{fan}}P_2 \tag{4}$$

where $\pi_{fan}$ is fan pressure ratio and $P_3$ is the fan exit air pressure and $P_2$ is the fan inlet air pressure. Also, fan output air temperature is calculated as follow;

$$T_3 = \frac{T_2}{\eta_{fan}}\left[\left(\frac{P_3}{P_2}\right)^{\frac{k_f-1}{k_f}} - 1\right] + T_2 \tag{5}$$

where $\eta_{fan}$ is the fan isentropic efficiency and $T_3$ is the fan exit air temperature and $T_2$ is the fan inlet air temperature and $K_f$ is the specific heat at constant pressure to specific heat in constant volume for the passing air, which is calculated as a function of the average temperature of the passing air. The exit air pressure of the high-pressure compressor (HPC) is obtained as follows;

$$P_{4=\pi_c} P_3 \tag{6}$$

where $\pi_c$ is compressor pressure ratio of HPC and $P_4$ is HPC output pressure and $P_3$ is the fan exit air pressure the exit temperature of the compressor is obtained as follow;

$$T_4 = \frac{T_3}{\eta_c}\left[\left(\frac{P_4}{P_3}\right)^{\frac{k_c-1}{k_c}} - 1\right] + T_3 \tag{7}$$

where $\eta_c$ is the compressor isentropic efficiency and $T_4$ is HPC output air temperature and $T_3$ is the fan output air temperature and $K_c$ is the specific heat at constant pressure to specific heat in constant volume for passing air of fan is calculated as a function of the average air passing temperature of the fan. Engine intake air real mass flow rate is calculated as follow;

$$m_{total} = \rho V_0 A \tag{8}$$

where A is a cross-section of an air inlet flow and $\rho$ is the density of inlet airflow and $V_0$ is flight velocity. The hot stream real air mass flow rate is obtained as follows;

$$m_{ah} = \frac{m_{total}}{\alpha + 1} \tag{9}$$

where $m_{total}$ is Engine intake air real mass flow rate and $m_{ah}$ is hot stream real air mass flow rate of engine core. Compressor power is obtained as follows;

$$W_C = m_{ah} C p_3 \frac{T_3}{\eta_c}\left[\left(\frac{P_4}{P_3}\right)^{\frac{k_c-1}{k_c}} - 1\right] \tag{10}$$

where $W_C$ is High-pressure compressor power consumption and $\eta_C$ is compressor isentropic efficiency and $K_c$ and $Cp_3$ are the specific heat at constant pressure to specific heat in constant volume and specific heat at constant pressure are calculated as a function of the average of compressor air temperature. $m_{ah}$ is hot stream air mass flow rate. fan power consumption is obtained as follows;

$$W_{fan} = m_{ac} Cp_2 \frac{T_2}{\eta_{fan}} \left[ \left(\frac{P_3}{P_2}\right)^{\frac{k_f-1}{k_f}} - 1 \right] \tag{11}$$

where $W_{fan}$ is fan power consumption and $\eta_{fan}$ is fan isentropic efficiency and $K_f$ and $Cp_2$ are the specific heat at constant pressure to specific heat in constant volume and specific heat at constant pressure are calculated as a function of the average fan air temperature. $m_{ac}$ is bypass air rate.

$$m_{ac} = m_{total} - m_{ah} \tag{12}$$

where $m_{ac}$ is bypass real mass flow rate and $m_{ah}$ is hot stream real air mass flow rate of engine core. Combustion chamber (CC) output pressure is obtained as follow;

$$P_5 = P_4 - \Delta P_{CC} \tag{13}$$

where $P_5$ is the CC output pressure, and $P_4$ is the compressor inlet pressure. Also, $\Delta P_{CC}$ is a pressure drop in the combustion chamber, it is evaluated as a percentage of the air exit pressure of the compressor. The heat transfer rate is calculated as follow;

$$Q_h = m_{ah} C_{av} (T_5 - T_4) \tag{14}$$

where, $Q_h$ is heat rate and $C_{av}$ is the specific heat at constant pressure is calculated as a function of average flow temperature in the combustion chamber (CC). also $T_4$ is inlet air temperature of turbine and $T_5$ is output air temperature of the high-pressure turbine (HPT). So, the fuel mass flow rate is calculated as follow;

$$m_f = \frac{Q_h}{FHV \, \eta_{CC}} \tag{15}$$

where FHV is fuel heat value per kilograms and $\eta_{CC}$ is combustion efficiency at the combustion chamber and $m_f$ is fuel consumption and $Q_h$ is heat rate. Turbine intake mass flow rate is calculated as follow;

$$m_T = m_{ah} + m_f \tag{16}$$

where $m_f$ is fuel consumption rate and $m_{ah}$ is hot stream real air mass flow rate. High-pressure turbine power is calculated as follow;

$$W_{HPT} = m_T \cdot C_{Pavt} \cdot (T_5 - T_6) \tag{17}$$

where, $m_T$ is turbine inlet mass flow rate and $C_{pav}$ is the specific heat at constant pressure which is obtained as a function of the average temperature in the high-pressure turbine. Low-pressure turbine power is calculated as follow;

$$W_{LPT} = m_T \cdot C_{Pavt} \cdot (T_6 - T_7) \tag{18}$$

where $m_T$ is turbine inlet mass flow rate and $C_{pav}$ is the specific heat at constant pressure which is obtained as a function of the average temperature in the Low-pressure turbine. Assuming that mechanical power losses in the spools are negligible, according to the energy conservation law, the High-pressure turbine power is equal to the High-pressure compressor power.

$$m_T \cdot C_{Pavt} \cdot (T_5 - T_6) = m_{ah} C p_3 \frac{T_3}{\eta_C} \left[ \left(\frac{P_4}{P_3}\right)^{\frac{k_c-1}{k_c}} - 1 \right] \tag{19}$$

where $T_6$ is the turbine exit temperature and $T_5$ is the HPT inlet temperature, $m_T$ is the turbine mass flow rate and $C_{pavt}$ is the specific heat at constant pressure, which is obtained as a function of the average temperature during the turbine. $\eta_c$ is also the compressor isentropic efficiency. Also, $k_c$ is the ratio of specific heat at constant pressure to the specific heat of the same volume as air, which is calculated as a function of the average air temperature during compressor. By obtaining the HPT output temperature of the equation (19), the turbine exit pressure is calculated as follows;

$$P_6 = P_5 \left[ 1 - \frac{1}{\eta_t} \left( 1 - \frac{T_6}{T_5} \right) \right]^{\frac{k_t}{k_t - 1}} \tag{20}$$

where $T_6$ is the output temperature of the turbine, $T_5$ is the inlet temperature of the turbine, and $P_6$ and $P_5$, respectively, are the turbine output pressure and turbine inlet pressure. kt is the ratio of specific heat at constant pressure to specific heat at constant volume, which is a function of the average temperature during the turbine, and $\eta_T$ is the turbine isentropic efficiency. Assuming that mechanical power losses in the spools are negligible, according to the energy conservation law, the Low-pressure turbine power is equal to the fan power.

$$m_T \cdot C_{Pavt} \cdot (T_6 - T_7) = m_{ac} Cp_2 \frac{T_2}{\eta_{fan}} \left[ \left(\frac{P_3}{P_2}\right)^{\frac{k_f-1}{k_f}} - 1 \right] \quad (21)$$

where $T_7$ is the turbine exit temperature and $T_6$ is the HPT inlet temperature, $m_T$ is the turbine mass flow rate and $C_{pavt}$ is the specific heat at constant pressure, which is obtained as a function of the average temperature during the turbine. $\eta_c$ is also the compressor isentropic efficiency. Also, $k_c$ is the ratio of specific heat at constant pressure to the specific heat of the same volume as air, which is calculated as a function of the average air temperature during compressor. The turbine exit pressure is calculated as follows;

$$P_7 = P_6 \left[ 1 - \frac{1}{\eta_t}\left(1 - \frac{T_7}{T_6}\right) \right]^{\frac{k_t}{k_t-1}} \quad (22)$$

where $T_7$ is the output temperature of the turbine, $T_6$ is the inlet temperature of the turbine, and $P_7$ and $P_6$, respectively, are the turbine output pressure and turbine inlet pressure. $k_t$ is the ratio of specific heat at constant pressure to specific heat at constant volume, which is a function of the average temperature during the turbine, and $\eta_T$ is the turbine isentropic efficiency. Regardless of the temperature and pressure changes in the bypass duct, the flow temperature at the mixer output is calculated in such a way;

$$T_8 = \frac{T_7 C_{P7} m_{ah} + (m_{ac} C_{P3} T_3)}{C_{P7}} \quad (23)$$

where $C_{P3}$ and $C_{P7}$, respectively are a specific heat at constant pressure and at the turbine output, which is calculated as a function of the flow temperature. $C_{P8}$ is the specific heat at constant pressure nozzle and $K_n$ is the ratio of specific heat at constant pressure to especial heat at constant volume, which is calculated as a function of the average flow temperature in the nozzle at the mixer output, where $T_7$ and $T_8$, respectively, are the flow temperature at the fan output and the turbine output and mixer output, and $m_{ac}$ is the cold stream real air mass flow rate and $m_{ah}$ is the hot stream real air mass flow rate. Assuming the ideal gas, the flow at the mixer outlet and the flow pressure at the mixer outlet ($P_8$) are calculated in such a way;

$$P_8 = \frac{1}{M_w} \rho R_u T_8 \quad (24)$$

where $R_u$ and $\rho$ are the global constant gas from the mass conservation law and airflow density, respectively. The output mixer mass flow rate ($m_8$) is the overall mass flow rates of air passing through the engine core, fuel flow rate, and cold stream real air mass flow rate;

$$m_8 = m_{ah} + m_{ac} + m_f \tag{25}$$

where $m_{ac}$, $m_{ah}$, and $m_f$ are respectively mass of air passing through the cold path, the air passing flow through the engine core, and the fuel mass flow rate. The nozzle output pressure is calculated in such a way;

$$P_9 = P_8 \left[1 - \frac{1}{\eta_n}\left(1 - \left(\frac{T_9}{T_8}\right)\right)\right]^{\frac{K_n}{K_n-1}} \tag{26}$$

Respectively, $T_9$ and $T_8$ are the exit temperature and inlet temperature of the nozzle and the $\eta_n$ is the nozzle isentropic efficiency and $P_8$ and $P_9$, respectively, is the inlet pressure and exit pressure of the nozzle and $K_n$ is the ratio of specific heat at constant pressure to specific heat at constant volume, which is calculated as a function of the average flow temperature at the nozzle. The velocity of the flow at the nozzle output ($V_9$) is calculated in such a way;

$$V_8 = \left(2\eta_n \frac{K_n}{K_n-1}\frac{R_u}{M_w} T_8 \left[1 - \left(\frac{P_9}{P_8}\right)^{\frac{K_n-1}{K_n}}\right]\right)^{0.5} \tag{27}$$

where $R_u$ and $M_w$ are gas global constant and molecular weight of combustion products and fuel, respectively. Also, the nozzle and $K_n$ are the ratio of specific heat at constant pressure to the specific heat of the same volume, which is calculated as a function of the average flow temperature in the nozzle. The thrust force of the engine is calculated in such a way[24];

$$F = m_8(V_9 - V_0) + A_9(P_9 - P_0) \tag{28}$$

where $P_0$ and $V_0$, respectively are the ambient pressure and flight speed, and $A_9$ is the area of the nozzle output section, and F is the Thrust force of the motor and $m_8$ is the mixer exit mass flow rate. The specific fuel consumption (TSFC) is equal to the ratio of fuel mass flow rate to Thrust force,[24]

$$TSFC = \frac{m_f}{F} \tag{29}$$

Thermal efficiency ($\eta_{th}$) is calculated in such a way;

$$\eta_{th} = \frac{m_8(V_9^2 - V_0^2)}{2m_f FHV} \tag{30}$$

where $m_8$ is the mixer output mass flow rate. Propulsive efficiency ($\eta_P$) is calculated in such a way[24];

$$\eta_P = \frac{V_0 F}{m_8(V_9^2 - V_0^2)} \tag{31}$$

where $m_7$ is the mass flow rate. Exergy (Ex) is known as entropy-free energy (maximum amount of work) in each system [25]. Two types of exergies are defined in the thermodynamic analysis of the power generation cycles. The physical exergy and the chemical exergy are these two types. The physical exergy demonstrated the maximum workability that can be extracted from fluid flow and the chemical exergy designates the maximum work that can be extracted from fuel flow. Physical exergy of all engine components is premeditated for both inlet or outlet of airflow as;

$$e = (S - S_0) - T_0(H - H_0) \tag{32}$$

where S, $S_0$ H, $H_0$, and $T_0$ are respectively the specific entropy of the fluid flow, the specific entropy of the dead state, the specific enthalpy of the fluid flow, the specific enthalpy of the dead state, and the ambient air temperature. The exergy flow rate is equal to the generate of the air mass flow rate in the air-specific exhaust.[25]

$$Ex = m_a e \tag{33}$$

where e is the airflow physical exergy. Fuel flow chemical exergy rate is premeditated as follows[26];

$$Ex_f = m_f e_x \tag{34}$$

where $Ex_f$, $m_f$, and Exf are respectively the specific chemical exergy of the fuel flow, the flow rate of the fuel consumed by the engine, and the rate of fuel flow chemical exergy. The hydrocarbon fuels specific chemical exergy is intended as follows[26];

$$Ex = FHV \left(1.04224 + \left(0.011925 \frac{a}{b}\right) - \left(\frac{0.042}{a}\right)\right) \tag{35}$$

where FHV, a, and b are respectively the fuel heat value, the number of carbon atoms in each fuel molecule, and hydrogen atoms in each fuel molecule. The overall exergitic efficiency of the turbofan engine is equal to the flight velocity ratio multiplied by the Thrust force to input fuel flow chemical exergy rate[26].

$$\eta_{ex} = \frac{F.v_o}{Ex} \tag{36}$$

where $\eta_{ex}$ is the overall exergetic efficiency. The fan exergy efficiency is intended as follows [26];

$$\eta_{exc} = \frac{E_3 - E_2}{W_f} \tag{37}$$

where $W_f$, $E_3$, and $E_2$ are respectively the fan mechanical power consumption, the fan exit exergy rate, and the fan inlet exergy rate. The fan exergy destruction rate ($E_{Df}$) is premeditated as[26];

$$E_{Df} = W_f + E_2 - E_3 \tag{38}$$

The compressor exergitic efficiency is intended as follows[26];

$$\eta_{exc} = \frac{E_4 - E_3}{W_c} \tag{39}$$

where $E_4$, $E_{ic}$, and $W_c$ are the compressor outlet flow exergy rate, the compressor inlet flow exergy rate, and the consumed mechanical power of the compressor, respectively. Correspondingly, the rate of exergy destruction in the compressor ($E_{DC}$) is calculated as follows[26];

$$E_{DC} = W_c + E_3 - E_4 \tag{40}$$

The High-pressure turbine exergy efficiency is calculated as[26];

$$\eta_{exHPT} = \frac{W_{HPT}}{E_5 - E_6} \tag{41}$$

where $W_{HPT}$, $E_5$, $\eta_{exHPT}$ and $E_6$ are respectively the turbine power output, the turbine exergy efficiency, the turbine input flow exergy rate, and the High-pressure turbine outlet flow exergy rate. Correspondingly, the turbine exergy destruction rate ($E_{DHPT}$) is premeditated as follows [26];

$$E_{DHPT} = E_5 - E_6 - W_{HPT} \tag{42}$$

Low-pressure turbine exergy efficiency is calculated as[26];

$$\eta_{exLPT} = \frac{W_{LPT}}{E_6 - E_7} \tag{43}$$

where $W_{LPT}$, $E_5$, $\eta_{exLPT}$ and $E_7$ are respectively the Low-pressure turbine power output, the turbine exergy efficiency, the Low-pressure turbine input flow exergy rate, and the turbine outlet flow exergy rate. Correspondingly, the Low-pressure turbine exergy destruction rate ($E_{DLPT}$) is premeditated as follows [26];

$$E_{DLPT} = E_6 - E_7 - W_{LPT} \tag{44}$$

Also, The combustor chamber exergy efficiency is intended in this way[26];

$$\eta_{exCC} = \frac{E_5}{E_4 - E_f} \tag{45}$$

where $E_f$, $\eta_{exCC}$, $E_4$, and $E_5$ are respectively the chemical fuel flow exergy rate, the combustor exergy efficiency, combustor input flow exergy rate, and output flow exergy rate. The exergy destruction rate is considered in the combustor chamber as ($E_{Dcc}$) as follows [26];

$$E_{Dcc} = E_4 - E_5 + E_f \tag{46}$$

Also, The Mixer exergy efficiency is intended in this way [26];

$$\eta_{exmixer} = \frac{E_8}{E_3 + E_7} \tag{47}$$

where $E_7$, $\eta_{exmixer}$, $E_3$, and $E_8$ are respectively the LPT output flow exergy rate, the mixer exergy efficiency, fan output flow exergy rate, and mixer output flow exergy rate. The exergy destruction rate is considered in the mixer as ($E_{Dmixer}$) as follows [26];

$$E_{Dmixer} = E_7 + E_3 - E_8 \tag{48}$$

Also, The fan exergy efficiency is intended in this way [26];

$$\eta_{exnozzle} = \frac{E_9}{E_8} \tag{49}$$

where $E_8$, $\eta_{exnozzler}$, $E_9$ are respectively the LPT output flow exergy rate, the nozzle exergy efficiency, fan output flow exergy rate, and nozzle output flow exergy rate. The exergy destruction rate is considered in the mixer as ($E_{Dnozzle}$) as follows [26];

$$E_{Dnozzle} = E_8 - E_9 \tag{50}$$

## 2.2. Deep learning

The artificial neural network is a significant part of machine learning techniques that can be found in a large set of data inspired by the performance of the brain. The artificial neural network is mainly composed of three parts: the first part consists of an input layer; the middle section contains one layer or more multi-hidden layers, and the last part contains an output layer, which is shown in Fig.2.

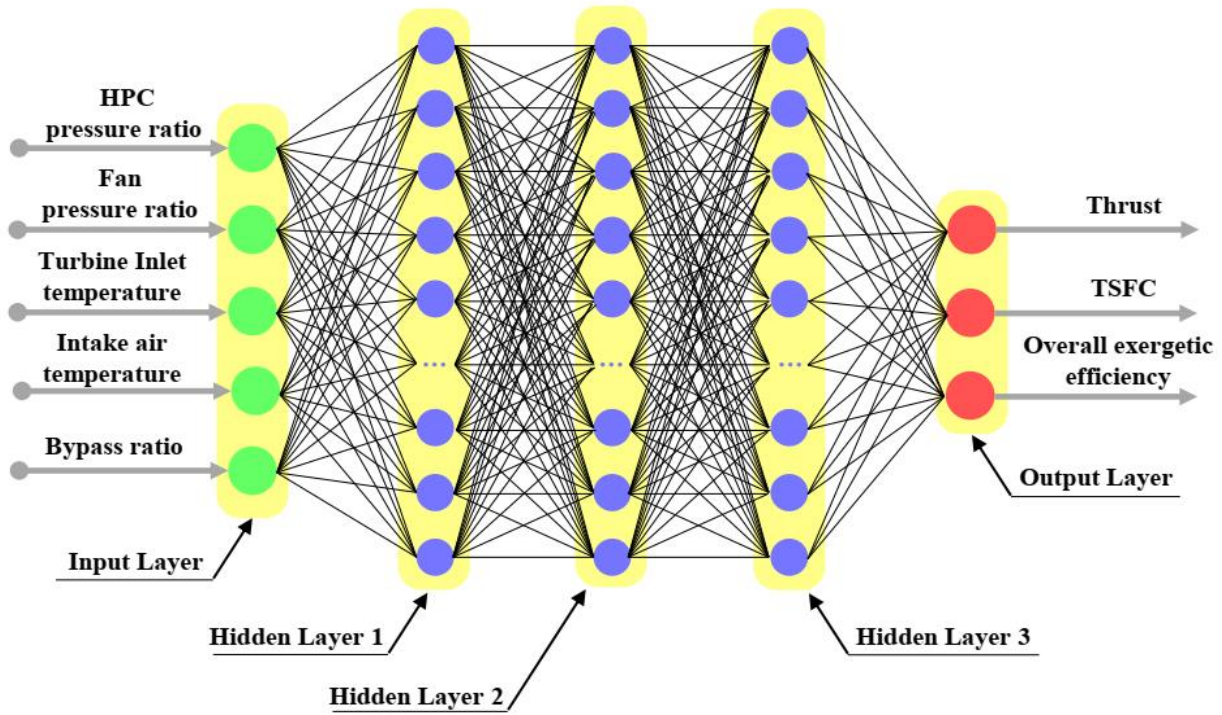

**Fig. 2.** schematic of the deep neural network.

Deep learning is a kind of neural network that has several layers and significant neurons in the hidden and output parts. The present study has received the network, the data, and the parameters affecting the process such as fan pressure ratio, high-pressure compressor (HPC) pressure ratio, variation of intake air temperature (IAT), turbine inlet temperature (TIT), and bypass ratio (alpha) in the input layer and gives the results in the output layer after deciphering the hidden layers.

As shown in Fig.2, hidden layer and output layers are composed of artificial neurons that have several weight coefficients and activation functions. The input values of each neuron are collected together after a weighting process, and then the output of the neuron is calculated by the activation function. Activation functions as an example can be selected from among the Tanh, sigmoid, and Rectified Linear Unit (RLU), which RLU has attracted much attention and are defined as follows;

$$f(x) = \begin{cases} x & if\ x > 0 \\ 0 & if\ x \leq 0 \end{cases} = \max(x, 0) \tag{51}$$

In the learning procedure of artificial neural networks, there are two basic approaches include supervised learning and unsupervised learning. The first one, labeled data will help to predict outcomes while another one does not do. In this study, such as many applications, we used a supervised one where the output is compared with the actual output. In particular, the weight of the neuron is calculated in such a way that the predicted output in return for a series of inputs can be as close as possible. It should be noted that quantification of network error for approximation of the results is done using cost function or loss function, one of the most common loss functions is Mean Squared Error (MSE) which is defined as follows;

$$L(y, \hat{y}) = \frac{1}{N} \sum_{i=1}^{N} (y_i - \hat{y}_i)^2 \tag{52}$$

where N is the number of samples, $y_i$ is the actual output, and $\hat{y}_i$ is the predicted output. The MSE function is often used for deep neural networks with the objective of approximation continuous values (regression problems). As mentioned, in the learning process, the weight calculation of each neuron is considered to minimize the loss function. First-order optimization algorithms based on the gradient descent method are one of the most popular methods for calculating the neuron's weight. If $w$ is the weight vector of the network weights, the loss function $L(w)$ can be defined as;

$$L(w) = \frac{1}{N} L_i(w) \tag{53}$$

where $L_i$ is the loss function of i-sampled and N is the number of the samples.

There are several optimization algorithms to minimize loss function such as standard gradient descent, stochastic Gradient Descent (SGD), Root Mean Square Propagation (RMSProp), and Adaptive Moment Optimization (ADAM), which the latter one has received a lot of attention due to computational efficiency, simplification in implementation, less memory requirement, etc.

Adam is considered as a combination of RMSProp and SGD methods with momentum because like RMSProp uses the square gradient to scale the learning rate and resembles the momentum benefits from the moving gradient mean. Adam calculates the momentum as the weighted average of the gradients as follows:

$$m_t^{(j)} = \beta_1 m_{t-1}^{(j)} + (1 - \beta_1) g_t^{(j)} \tag{54}$$

$$v_t^{(j)} = \beta_2 v_{t-1}^{(j)} + (1 - \beta_2)(g_t^{(j)})^2 \tag{55}$$

where $m_t$ is the first moment, $v_t$ is the second moment, and g is the gradient on the current mini-batch and also $\beta_1$ and $\beta_2$ are exponential decay rates respectively considered as 0.9 and 0.999. Then, the first moment and the second moment are bias-corrected as:

$$\widehat{m}_t^{(j)} = \frac{m_t^{(j)}}{1 - \beta_1^t}, \widehat{v}_t^{(j)} = \frac{v_t^{(j)}}{1 - \beta_2^t} \tag{56}$$

Also, the updating of the weights is carried out according to the equations (57) with alpha learning which is assumed to be equal to 0.001.

$$w_{t+1}^{(j)} = w_t^{(j)} - \frac{\alpha}{\sqrt{\widehat{v}_t^{(j)}} + \varepsilon} \widehat{m}_t^{(j)} \tag{57}$$

Using momentum instead of gradients to update neuron weights helps ADAM accelerate to find local minima.

*2.3. KERAS*

Keras is an application program interface designed not only for machines but also for humans. Keras is an open-source programming platform that builds simple and compatible APIs, reduces the number of user actions, and provides practical and clear error messages. Keras can run various machine learning libraries such as Theano (developed by the University of Montreal), CNTK (designed by Microsoft), and TensorFlow (acquired by Google). Keras, which provides fast and easy prototyping with features of modularity, scalability, and user-friendliness, has been ranked second after TensorFlow in 2018 in the power ranking, which emphasizes popularity, interest, and use. has taken. In the Keras platform, which is programmed in Python itself, neural networks with different structures can be created. Also, in Keras, all activation functions, loss functions, and even optimization algorithms such as Adam, RMSProp and SGD are applicable. As a result, due to its widespread use and unique benefits, Keras has also been used in the Python context in this study.

## 3. Validation

The F135 PW100 turbofan engine is a mixed-flow turbofan. This engine is used as a propulsion system of the Lockhid Martin F-35 Lightning. The components of the F135 PW100 engine include an axial-flow fan with three stages, a high-pressure compressor (HPC) with six stages, a combustion chamber (CC), a high-pressure turbine (HPT) with one stage, a low-pressure turbine (LPT) with two stages, mixer and nozzle [27]. The Input Parameters of the F135 PW100 engine are shown in Table 1 [27].

**Table 1.** Input parameters of F135PW100 engine modeling

| Parameters | Symbol | Value | Unit |
|---|---|---|---|
| Inlet air mass flow rate | $m_a$ | 147 | Kg/s |
| Fan pressure ratio | $\pi_{fan}$ | 4.7 | - |
| High-Compressor pressure ratio | $\pi_c$ | 6 | - |
| Isentropic efficiency of the fan | $\eta_{exc}$ | 0.90 | - |
| Isentropic efficiency of the high-pressure compressor | $\eta_{exHPC}$ | 0.85 | |
| High-pressure turbine inlet temperature | $TIT$ | 2175 | K |
| Isentropic efficiency of the high-pressure turbine | $\eta_{exHPT}$ | 0.90 | - |
| Isentropic efficiency of the low-pressure turbine | $\eta_{exLPT}$ | 0.91 | - |
| Isentropic efficiency of the combustion chamber | $\eta_{exCC}$ | 0.995 | - |
| Bypass ratio | $\alpha$ | 0.57 | - |

The efficacy of several fuel types such as hydrogen fuel, natural gas (LNG) has also been inspected on the performance of the proposed engine. The fuel heat value (FHV) and special chemical exergy of these fuels are given in Table 2. In this section, the efficiency of fuel types, flight altitude, and intake temperature variation on the F135 PW100 mixed-flow turbofan engine is investigated. First, the validation of the results obtained from the modeling in the take-off conditions (Ma = 0, and H = 0). The validation was evaluated for the thrust, the fuel consumption mass flow rate, and TSFC.

**Table 2.** The fuel heat value and special chemical exergy of fuels [28]

| Fuel type | JP10 | Diesel | Natural gas | Hydrogen |
|---|---|---|---|---|
| Special Chemical Exergy of Fuels (MJ/Kg) | 44.921 | 44.661 | 55.168 | 134.778 |
| Fuel Heat Value (MJ/Kg) [28] | 42.075 | 42.740 | 49.736 | 118.429 |
| Molecular Weight(g/mol) [28] | 136 | 167 | 16 | 2 |
| Chemical Formula[28] | $C_{10}H_{16}$ | $C_{12}H_{23}$ | $CH_4$ | $H_2$ |

The results of modeling in the take-off condition compared with reference values[27], also the validation outputs are represented in table 3. The consequences demonstrated that the results have an acceptable error for modeling results, which is verified by modeling accuracy.

**Table 3.** The fuel heat value and special chemical exergy of fuels

|  | **Modeling results** | **reference values**[27] | **Error** |
|---|---|---|---|
| Thrust (KN) | 118.580 | 125.903 | -5.81% |
| TSFC(g/KNs) | 26.43 | 25 | 5.72% |
| Fuel mass flow rate (Kg/s) | 3.134 | 3.15 | -0.5% |

## 4. Results and discussion

*4.1. Engine performance analysis*

*4.1.1. effect of Mach number and flight altitude on engine performance*

In this section, the impact of Mach number and flight altitude on the Intake air Mass Flow rate, TSFC, Thrust, propulsive efficiency, and thermal efficiency of the F135 PW100 engine was investigated. Also, in this case, JP10 is used as fuel.

The changes in Intake Air Mass Flow rate of the F135 PW100 engine with Flight-Mach number and flight altitude are presented in Fig. 3a. It has been observed that the flight velocity increases at each constant flight height with increasing Mach number. So, the intake air mass flow rate to the engine increases also, in each constant flight-Mach number, with increasing flight altitude, the intake mass flow rate to the engine decreases because the intake air density is decreased.

The variation of the F135 PW100 engine Thrust force with Mach number and flight altitude is represented by the use of JP10 fuel in Fig. 3b. At every constant flight height, with an increase in the flight-Mach number, intake airflow to the engine is increased, so the Thrust Force is increased. Also, the flight height of the input air to the Engine increases at constant the flight-Mach number, so the Thrust is reduced. Because, with an increase in the flight altitude, intake airflow to the engine is decreased.

The variations of the Thrust specific force (TSF) of the F135 engine are discussed at an altitude of 20,000 meters with flight-Mach using JP10 fuel. The modifications of the TSF of the F135 PW100 engine with flight altitude and flight-Mach number have been displayed in Fig. 3c by using JP10 fuel. The intake mass flow rate to the engine increases with increasing Flight-Mach Number at each constant flight height. Since with increasing flight Mach number, the intensity of the rising intake mass flow rate to the engine is higher than the intensity of the increase in the Thrust, so TSF decreases with the increasing Mach number.

The TSFC changes of the F135 PW100 engine with flight height and Mach number have been displayed using JP10 fuel in Fig. 3d. By increasing the intake air mass flow rate to the engine, so more energy is needed to reach the combustion chamber inlet temperature. Therefore, the heating rate and fuel consumption mass flow rate increase with increasing the inlet air mass flow rate.

The inlet air flow rate increases with Flight-Mach Number at each constant altitude, but as the intensity of the rise in the amount of fuel consumption mass flow rate is higher than that intensity of the amount of Thrust increasing, the TSFC increases with increasing Flight-Mach number due to the increase in the intake air mass flow.

Changes in thermal efficiency of F135 PW100 with Flight-Mach number and flight altitude are shown in Fig. 3e. The representation showed that at each constant flight height with increasing Flight-Mach number in the range of 1 to 2, thermal efficiency increases, and with increasing Flight-Mach number in the range of 2 to 2.5, thermal efficiency decreases.

The variation of the propulsive efficiency is represented with Flight-Mach number and altitude by using the JP10 fuel in Fig. 3f. Since the flight-Mach number increases, both the Thrust of the F135 engine and flight velocity increase, the propulsive efficiency increases with increasing Flight - Mach number in each constant flight altitude.

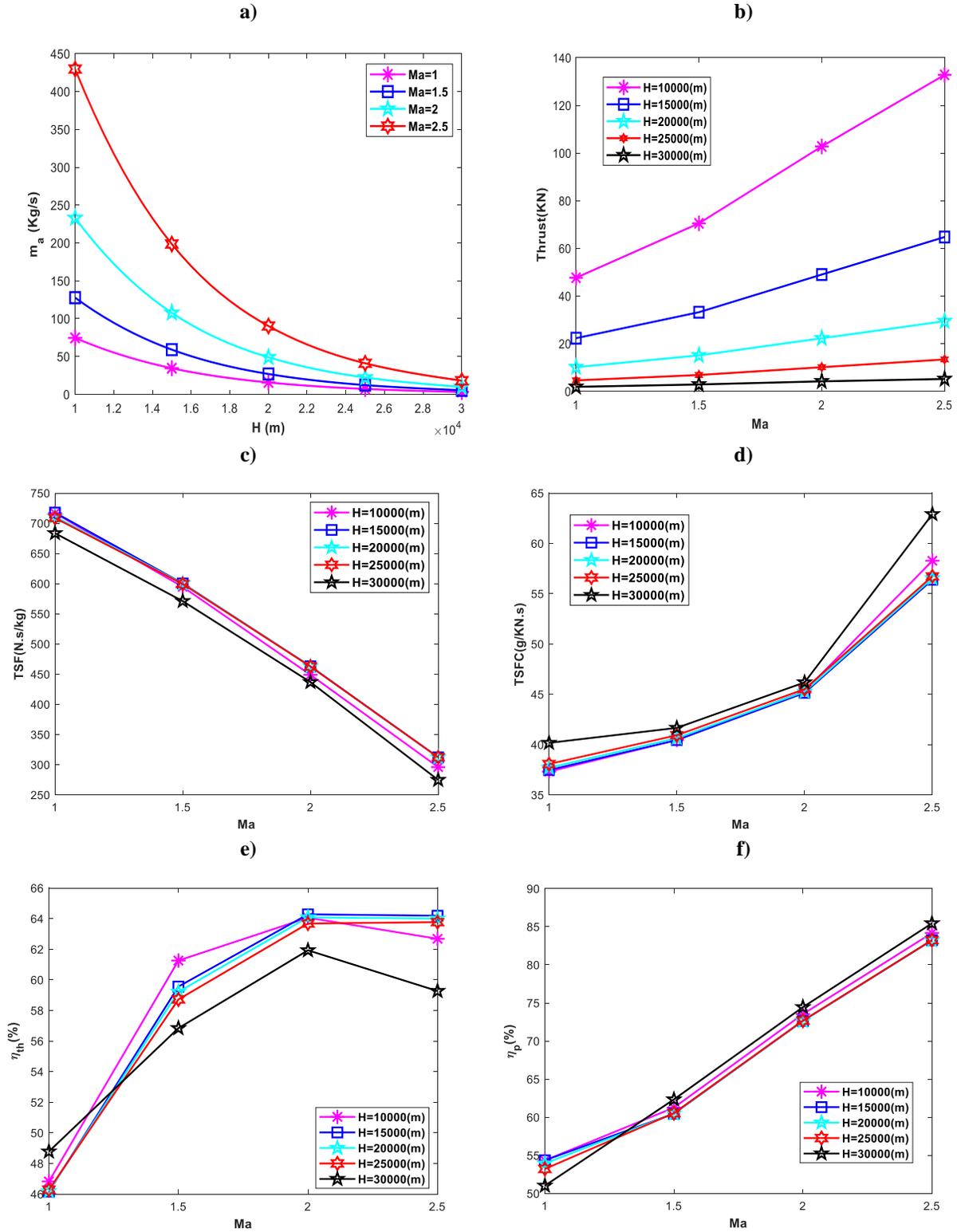

**Fig. 3.** the effects of Flight- Mach number and flight altitude of the F135 PW100 engine in a) the intake air mass flow rate b) net Thrust c) TSF d) TSFC e) thermal efficiency f) Thrust by using JP10 as fuel

*4.1.2. Effect of fuel types on engine performance*

Also, in this section, the effect of the use of different fuels, including natural gas and hydrogen fuel, diesel fuel, and JP10 on performance parameters such as Thrust, TSFC, thermal efficiency, and propulsive efficiency were investigated at the height of flight condition of 30,000 m altitude and Mach number of 2.5. Thrust force and TSFC changes are represented in Fig. 4a and Fig. 4b, respectively. The changes are applied by the fuel types at an altitude of 30,000 m and Mach number 2.5. Whatever, the molecular fuel weight is less, the molecular weight of combustion productions mixture of fuel with air is similarly less. The reduction of the molecular fuel weight increases the nozzle exit velocity. As a result, the velocity term of the Thrust force increases. So, the thermodynamic cycle has the highest Thrust force by using hydrogen as fuel, as shown in Fig. 4a.

Also, the TSFC changes of the F135 PW100 engine with fuel type at a flight height of 30,000m and Flight-Mach number of 2.5 are shown in Fig.4b. Consequently, the fuel consumption mass flow rate is reduced by increasing the fuel heat value (FHV). So, TSFC is reduced by increasing the FHV. Also, the thermal, propulsive, and overall efficiencies of the F135 engine are represented by the fuel type at a flight altitude of 30,000 m and Mach 2.5 in Fig. 4c. Since the reduction of the molecular fuel weight increases nozzle exit velocity, and Thrust force, the rate of increase in the kinetic energy of the flow along the engine increases with the reduction in the fuel molecular weight. Since the intensity of the increase of the thrust force due to the reduction of molecular fuel weight is lower than the intensity of the increase in the kinetic energy variations during the engine due to the reduction of molecular fuel weight, so, the propulsive efficiency decreases with decreasing the molecular fuel weight. Also, cause of the increase in the molecular weight of the fuel. The exit nozzle velocity is reduced. And the rate of variation of the kinetic energy of flow during the engine is reduced by increasing the molecular fuel weight. Consequently, thermal efficiency decreases with increasing the molecular fuel weight.

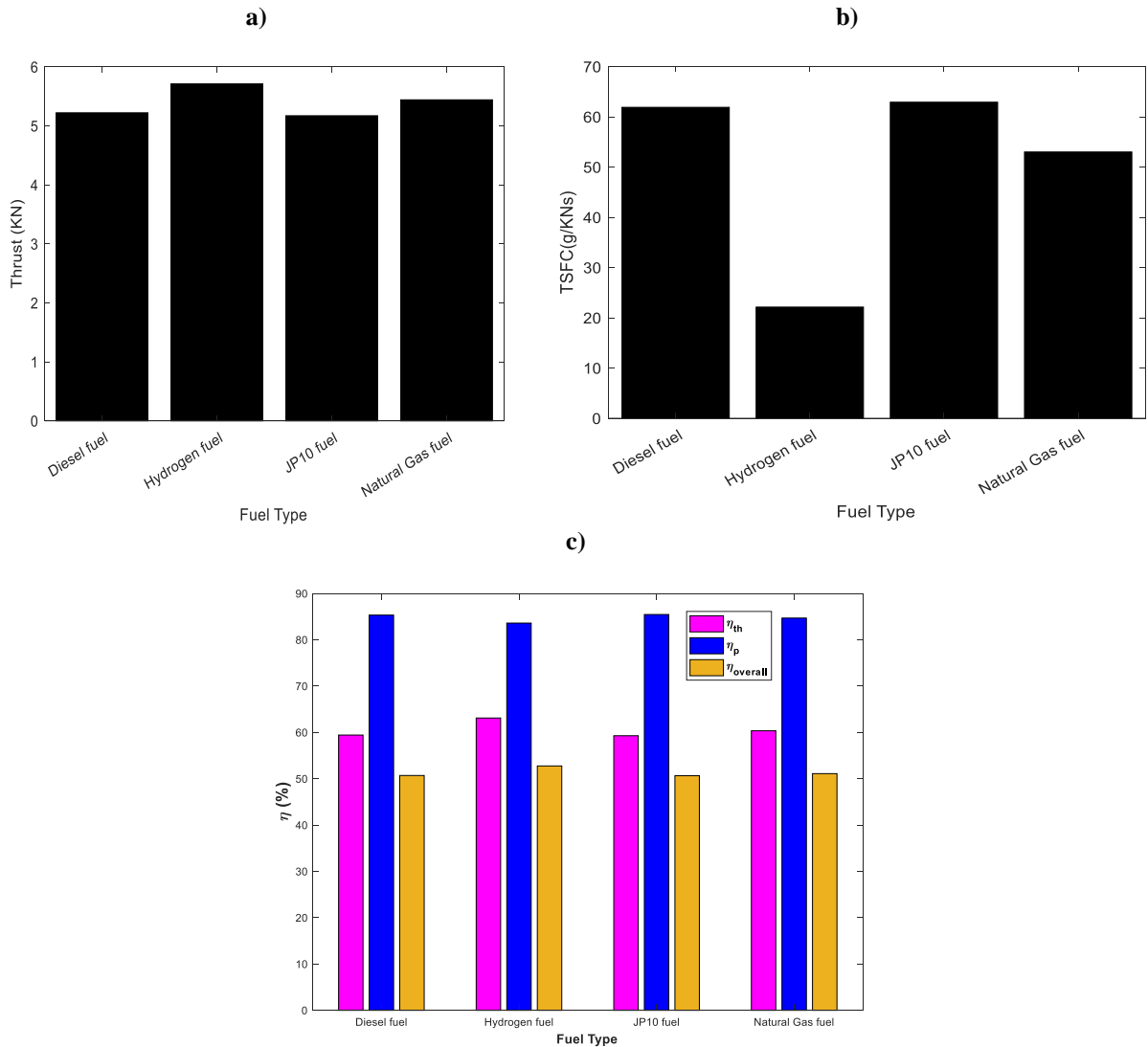

**Fig. 4.** the efficacy fuel type at an altitude of 30000m and Flight-Mach number of 2.5 on a) TSFC b) Thrust force c) thermal efficiency, propulsive efficiency, and overall efficiency

*4.1.3. Effect of inlet air temperature changes on engine performance*

In this section, the effect of intake air temperature variation on the Thrust, TSFC, thermal efficiency, and propulsive efficiency of the F135 PW100 turbofan engine is investigated by using hydrogen fuel at different flight altitudes and Flight-Mach numbers. Two fragments are disused,

Flight-Mach number is assumed as constant with the amount of 2 and flight altitude in the range of 10,000m to 30,000m, then the flight altitude is supposed constant as 20,000 m with varied Flight-Mach number. These fragments are applied at three parameters.

First, the changes in inlet air mass flow rate of F135 PW100 in terms of difference inlet air temperature with different flight altitudes and Flight-Mach numbers have shown in Fig. 5a and Fig. 5b. It has been observed at each flight altitude and Mach number, with decreasing inlet air temperature, the inlet air density increases, so the inlet air mass flow rate is increased.Second, the changes in the Thrust of the 135 PW100 engine with inlet air temperature are represented in Fig. 5c and Fig. 5d at different flight conditions. the intake air mass flow rate increases with decreasing the inlet air temperature in each flight condition . So, the Thrust is increased.

Third, TSFC variation of the F135 PW100 engine is represented by decreasing inlet air temperature at different flight-Mach numbers and altitudes in Fig. 5e and Fig. 5f. In each case, reducing the intake air temperature requires more energy to deliver the flow temperature to the limit of the turbine inlet temperature (TIT) so the consumption fuel mass flow rate is increased. Because the combustion chamber inlet temperature is reduced in terms of the engine inlet temperature drop.

But because the intensity of the increase in the fuel consumption mass flow rate by decreasing the intake air temperature is higher than the intensity of Thrust force increasing by reducing the intake air temperature, accordingly, TSFC is increased by reducing the intake air temperature at a flight altitude of the 20,000 m in the Mach number ranges under Ma=1. however, in the Mach number above 1, because the intensity of the increase in the fuel consumption mass flow rate with intake air temperature decrease is lower than the intensity of Thrust increasing with intake air temperature reducing, so, TSFC is decreased by intake air temperature reducing at Flight-Mach number of 2 and flight altitude in the range 10,000 m to 30,000 m. Also, the applied fragment is considered at the other performance parameters such as thermal efficiency, propulsion efficiency.

The thermal efficiency variation of the F135 PW100 engine with decreasing intake air temperature at different Flight-Mach numbers and flight altitudes is shown in Fig. 6a and Fig. 6b. In each flight case, the air mass flow rate is increased by decreasing intake air temperature. Therefore, the variation rate of increasing the kinetic energy of airflow along with the engine increases. Since the intensity of the increase in variation rate of the kinetic energy during the engine is higher than the intensity of the increase in heating rate due to the decreasing intake air temperature. So, the thermal efficiency increased with decreasing intake air temperature. Correspondingly, the change of propulsive efficiency is represented by reducing the temperature of the inlet air at Flight-Mach number and the different heights in Fig. 6c and Fig. 6d.

The representation showed that the kinetic energy of airflow is increased by reducing the difference temperature of intake flow along with the engine. However, since the intensity rate of increase in the kinetic energy along with the engine is higher than the intensity of the increase in the thrust force due to the reduction in the intake air temperature, hence the input air temperature is reduced in any state, so it is the cause of decreasing in propulsive efficiency.

*4.2. Exergy analysis results*

*4.2.1. Effect of Flight altitude and Flight-Mach number*

In this section, the exergy efficiency, and the exergy destruction rate of F135 engine components containing the fan, high-pressure compressor (HPC), high-pressure turbine (HPT), the low-pressure turbine (LPT), nozzle, and the mixer are investigated by using JP10 as a fuel. In addition, the exergy efficiency and exergy destruction rate of the F135 PW100 engine with Mach number and altitude were analyzed by using JP10 as a fuel. The exergy efficiency and exergy destruction rate changes of the F135 PW100 engine and its components with changing the Flight-Mach numbers and flight altitudes are indicated in Fig. 7a-d.

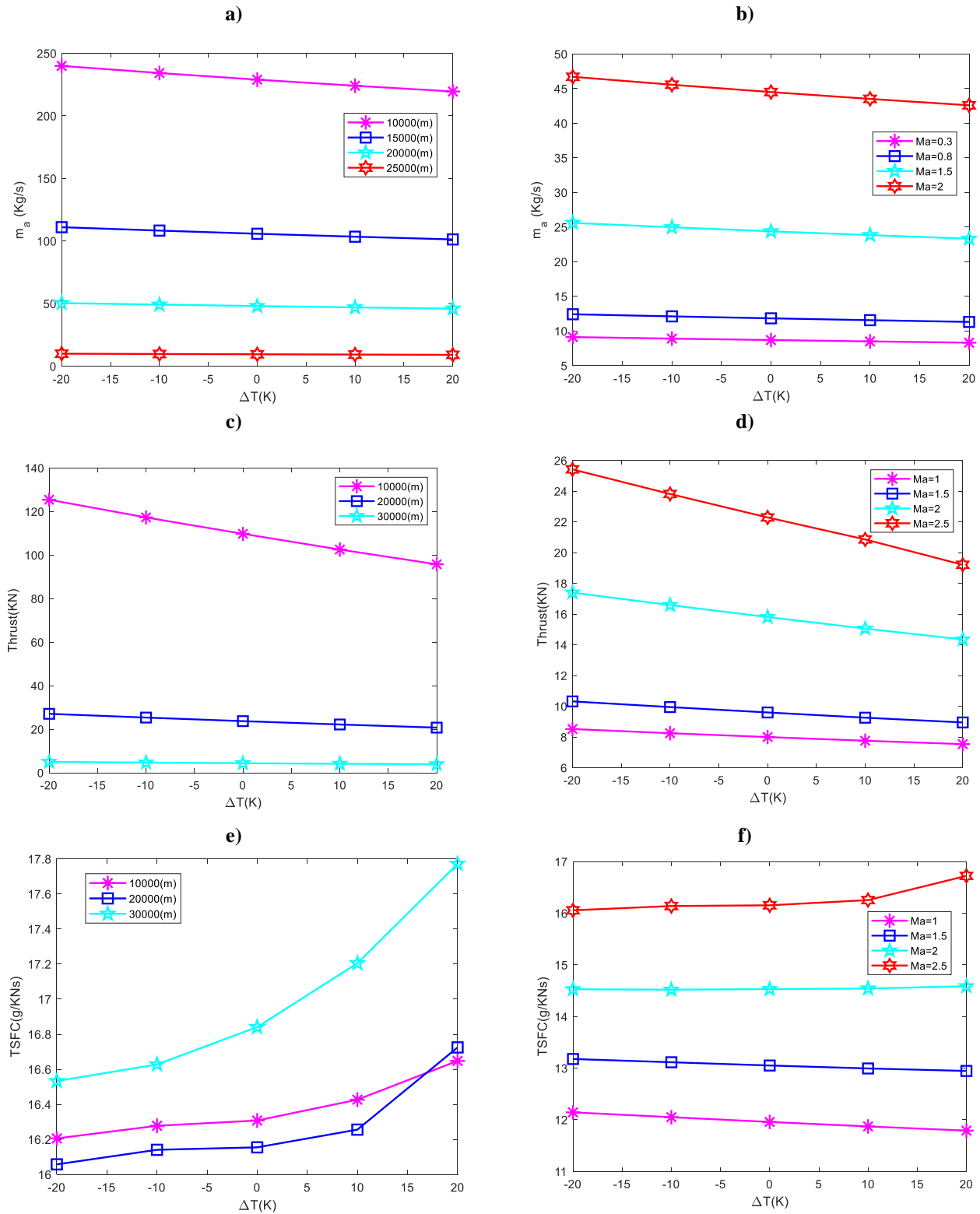

**Fig. 5.** The influences of difference inlet air temperature of the F135 PW100 engine with a) intake mass flow rate-varied flight altitude b) intake mass flow rate-varied Flight-Mach number c) Thrust-varied flight altitude d) Thrust -varied Flight-Mach number e) TSFC-varied flight altitude d) TSFC -varied Flight-Mach number

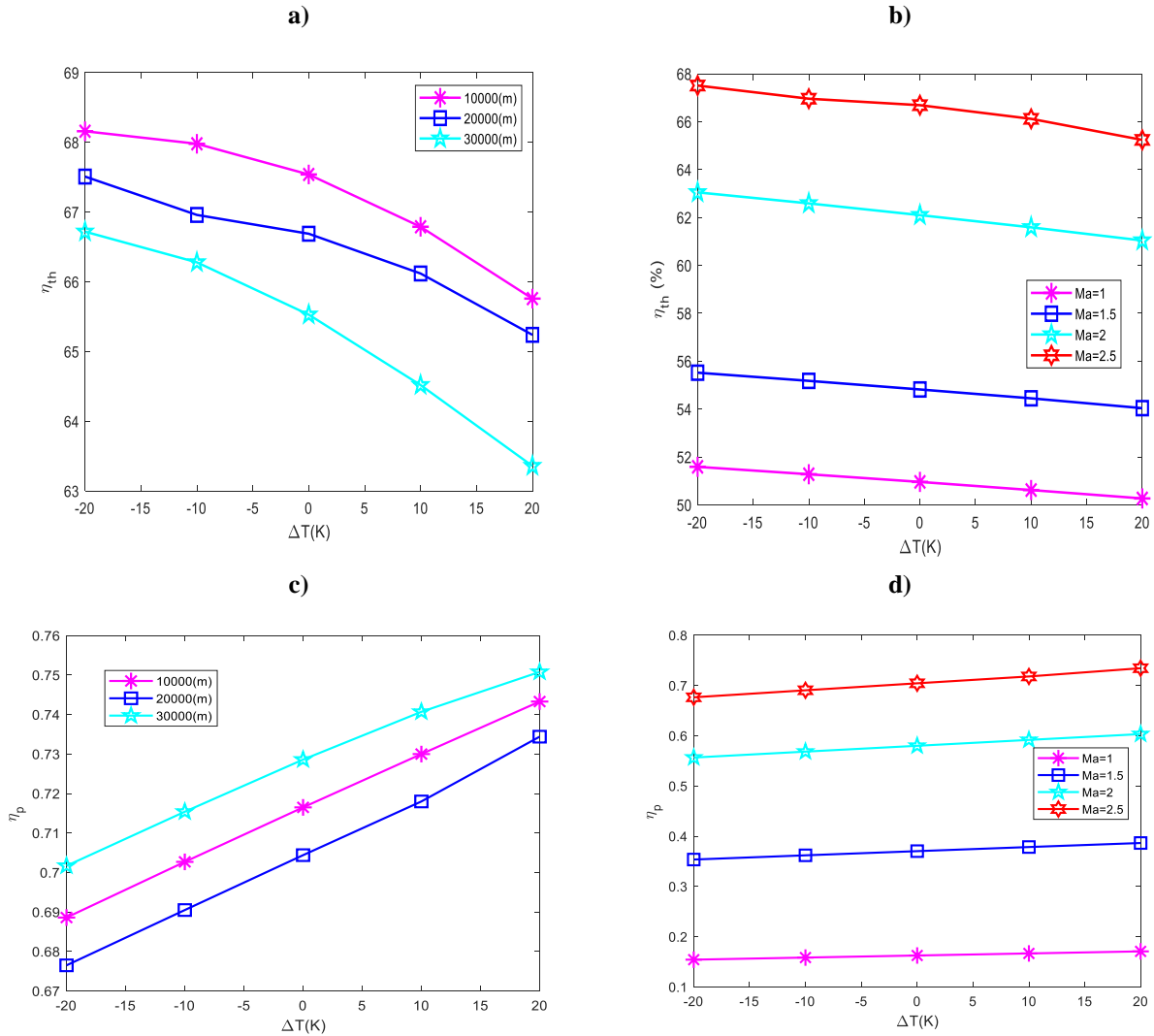

**Fig. 6.** The effects of difference inlet air temperature of the F135 PW100 engine in terms of a) Flight altitude variation in the range of 10,000 to 30,000 and constant Flight-Mach number of 2 for thermal efficiency b) Flight-Mach number Variation and constant flight altitude of 20,000 m for thermal efficiency c) Flight altitude variation in the range of 10,000 to 30,000 and constant Flight-Mach number of 2 for propulsive efficiency d) Flight-Mach number Variation and constant flight altitude of 20,000 m for propulsive efficiency

In Fig. 7a and Fig. 7c, the flight-Mach number is assumed as constant 2, and also, at Fig. 7b and Fig. 7d, the flight altitude is presumed as constant 15000 m. Both, the mixer exergitic efficiency and the overall exergitic efficiency were increased by increasing Flight-Mach number at the constant flight altitude. Also, the variation of the exergy destruction rate of the F135 PW100 engine and its components with Mach number and flight height is shown in Fig. 7c and Fig. 7d, the results exhibited that the exergy destruction rate of the components and overall exergy destruction rate of the F135 PW100 engine is decreased with increasing the flight altitude at constant Flight-Mach number. Also, the exergy destruction rate of each component is increased at the constant flight altitude by increasing the Flight-Mach number.

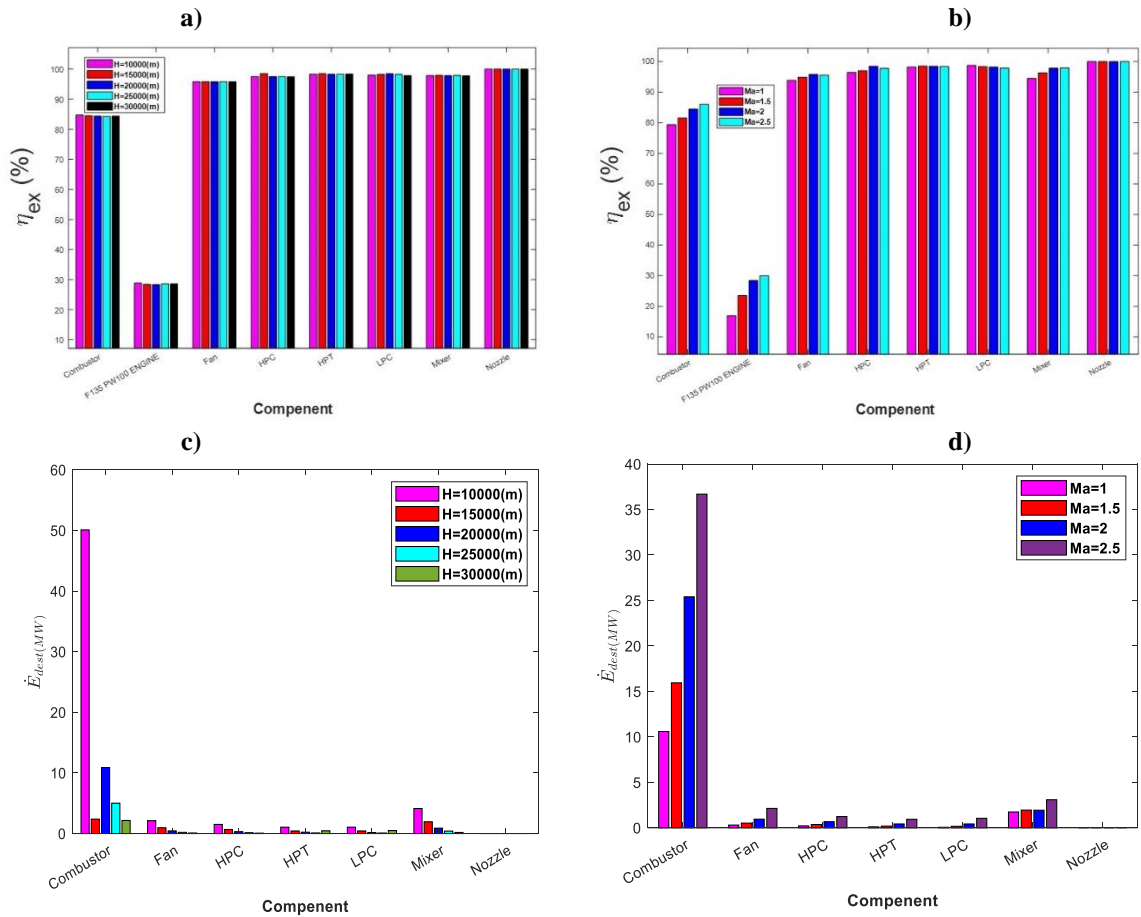

**Fig. 7.** efficacy flight altitude and flight-Mach number of the F135 PW100 engine and its components a) exergetic efficiency with the constant flight-Mach number of 2 b) exergetic efficiency with a constant flight altitude of 15000 m c) exergy destruction with constant flight-Mach number of 2 d) exergy destruction with a constant flight altitude of 15000 m

*4.2.2. Effect of inlet air temperature changes*

In this section, the effect of reducing the intake air temperature on the overall exergy efficiency of the F135 PW100 engine and F135 PW100's components including, the high-pressure compressor (HPC), high-pressure turbine (HPT), low-pressure turbine (LPT), fan, and mixer was investigated at an altitude of 20,000 m and Mach number of 2 by using JP10 as a fuel.

The overall exergy efficiency changes of the F135 PW100 engine and its components with intake air temperature are indicated in Fig. 8a. The results evidenced that the overall exergy efficiency of F135 PW100 and exergy efficiency of HPT, LPT, HPC, and fan are increased by decreasing the intake air temperature, and exergy efficiency of the combustion chamber and mixer is reduced by reducing the intake air temperature. Also, the changes of the overall exergy destruction rate of the F135 PW100 engine and components of F135 PW100 are indicated in Fig. 8b with intake air temperature changing at flight conditions of 20,000 m altitude and Mach number 2 by using JP10 as a fuel. The results confirmed that the exergy destruction rate of Fan, HPT, LPT, and HPC of F135 PW100 is reduced by reducing the intake air temperature. Consequently, the exergy destruction rate of the combustion chamber and mixer is increased.

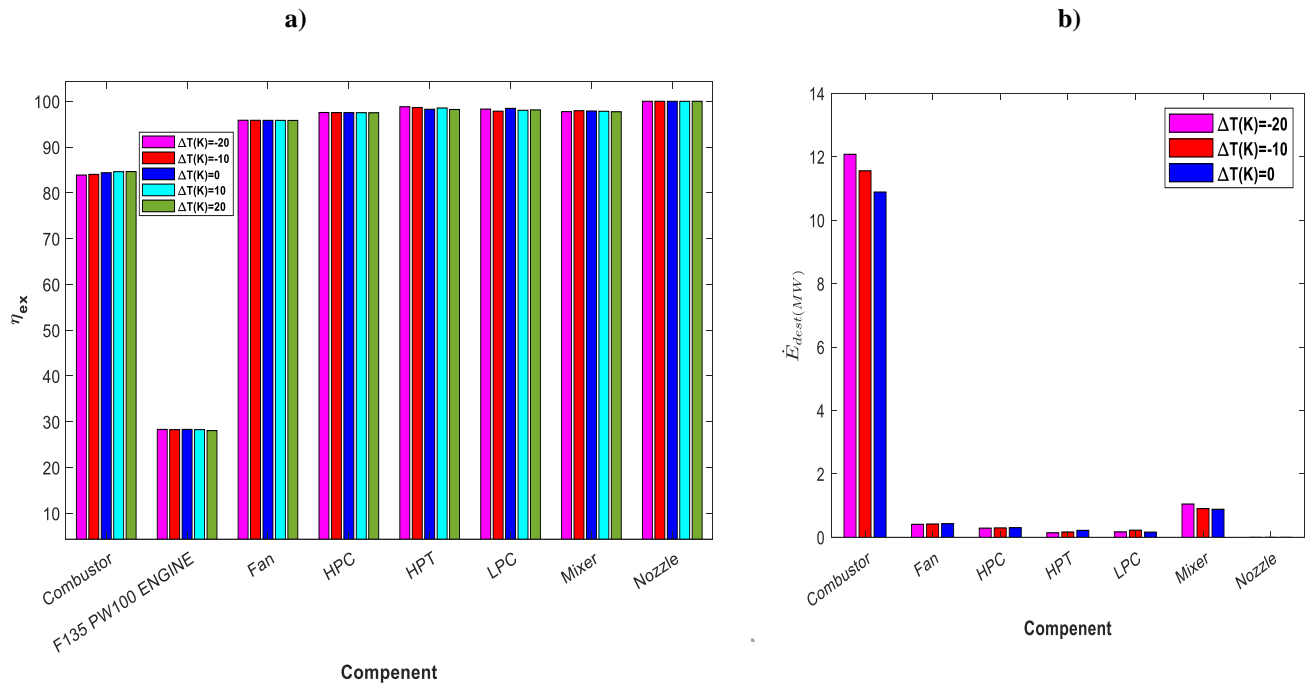

**Fig. 8.** the effects of difference inlet air temperature of the F135 PW100 engine and its components engine at a flight altitude of 20000 m and Flight-Mach number of 2 via using JP10 fuel on a) exergetic efficiency b) exergy destruction

*4.2.2. Effect of fuel types*

In this section, the exergy efficiency and exergy destruction rate of the F135 PW100 engine and its components of it including the high-pressure compressor (HPC), the high-pressure turbine (HPT), low-pressure turbine (LPT), and mixer at an altitude of 20,000 m and Flight-Mach number of 2 are investigated by using Hydrogen, Natural Gas and JP10 as fuels.

The exergy efficiency changes of the F135 PW100 engine and its components with the fuel type used at 20000m altitude and Flight-Mach number of 2 are shown in Fig. 9a. The results demonstrated that among the fuels investigated, the JP10 has the maximum overall exergy efficiency of the F135 PW100 engine, and the highest exergy efficiency is produced by the nozzle. Also, Hydrogen fuel has the lowest overall exergy efficiency of the F135 PW100 engine and the lowest exergy efficiency achieved at the Combustor chamber.

Correspondingly, the effects of the exergy destruction rate of the F135 PW100 engine cycle and F135 PW100 engine's components with the different fuel types used at 20000 m altitude and Mach number of 2 are shown in Fig. 9b. The results demonstrated that the least exergy destruction rate is achieved by a nozzle in the case of JP10 fuel use. Also, the highest exergy destruction rate is produced by the combustion chamber in the case of JP10 fuel use.

In the application of hydrogen fuel, the highest exergy destruction rate has occurred in all of the F135 PW100 engine's components. Also, the application of JP10 has the lowest exergy destruction rate in all of the components of the F135 PW100 engine.

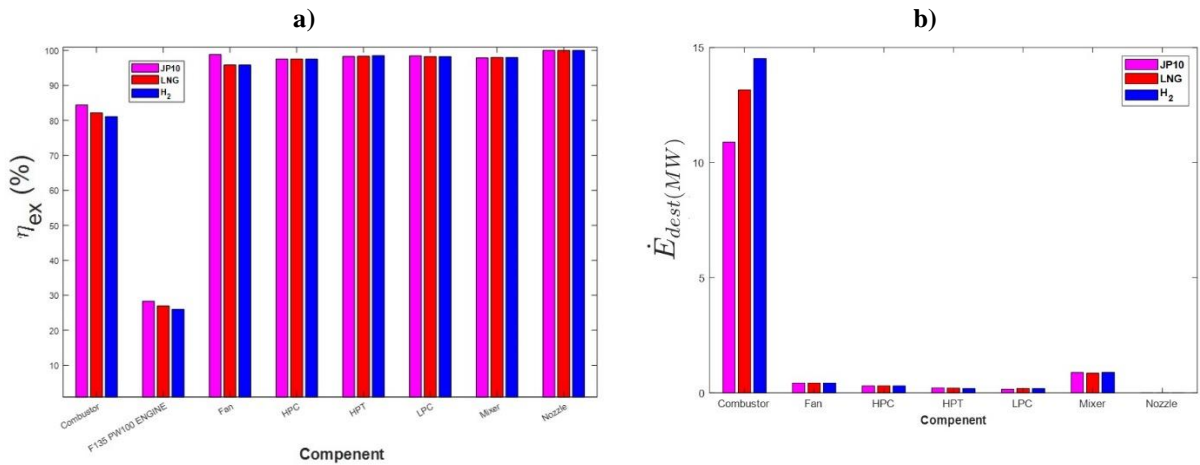

**Fig. 9.** the efficacy fuel types of the F135 PW100 engine and its components at a flight altitude of 20000 m and Flight-Mach number of 2 on a) exergetic efficiency b) exergy destruction

*4.3. Deep learning results*

Based on the results obtained in the subsections of 4.1, and 4.2, to model the thermodynamic performance of the F135 PW100 engine cycle, flight-Mach number and flight altitude are considered to be 2.5 and 30,000 m, respectively, due to the operational advantage of flying at ultrasonic altitude, and higher trust of hydrogen fuel. Accordingly, the appropriate datasets are provided and then randomly divided into two separate sets: the first set contains 6079 samples for model training and the second set contains 1520 samples for testing. Figure 10 shows the relationship between all inputs and outputs. In the present deep neural network model, five input variables including HPC, LPC, TIT, IAT, and bypass ratio (alpha) are considered as the characteristics of the model and three output variables consist of thrust, TSFC, the exergetic efficiency as labels. The data were extracted from python 3.9 software. To facilitate the training process, the input variables are normalized between zero and one. In this study, the Adam optimization algorithm, the cost function of the MSE, and the active function of Relu are used to train the network. The final deep network prototypical, as shown in Fig. 10, has three hidden layers, the first to third layers have 512, 256, and 128 neurons, respectively, which are manually adjusted based on experience.

The loss function diagram in terms of each epoch is shown in Figure 11 for each of the outputs such as thrust, TSFC, and exergetic efficiency. The results show a stable convergence process that has reached a very small amount for each output. The convergence trend is similar for all three outputs, although the final loss value of exergetic efficiency is greater than the TSFC, and for the TSFC is greater than the thrust. This is due to the difference in the level of the output values that are observed in the figure (see the scales of outputs parameters such as thrust, TSFC, and exergetic efficiency) . It is also observed that for each of the outputs, on average in the first ten epochs, the loss function is reduced by 90%.

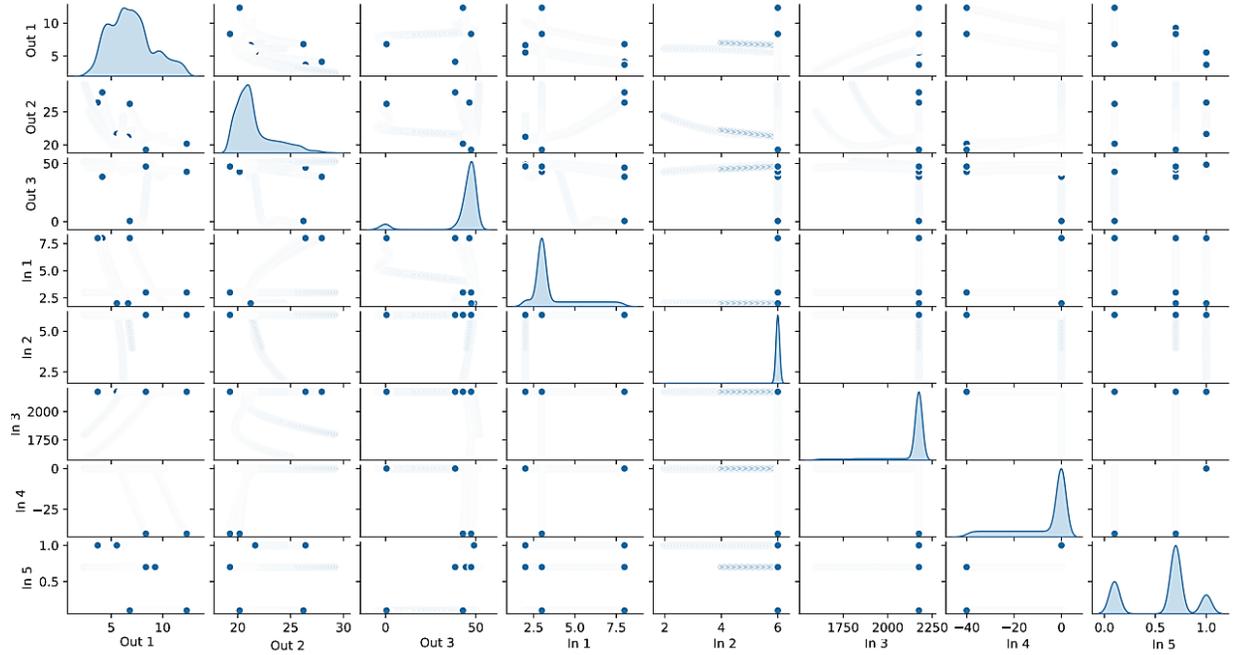

**Fig. 10.** The general figures of input and output variable's relationships

To show the performance of the deep learning method in predicting outputs, Figure 12 shows the predicted value of Thrust, TSFC, and exergetic efficiency with respect to the True values in the test section. It can be seen that the deep neural model has been very successful in predicting outputs; Because the blue points inside the diagrams are very close to the midline $P = T$. This means that the predicted value is very close to the corresponding true value. In some of these diagrams, it can be seen that the data aggregation and also the distance of some of them from the centerline is greater, which indicates the greater numbers of data in these areas and, consequently, the greater the probability of error. For example, to predict the second output (TSFC), the accumulation of blue dots is greater in the range $21 \leq TSFC \leq 22$.

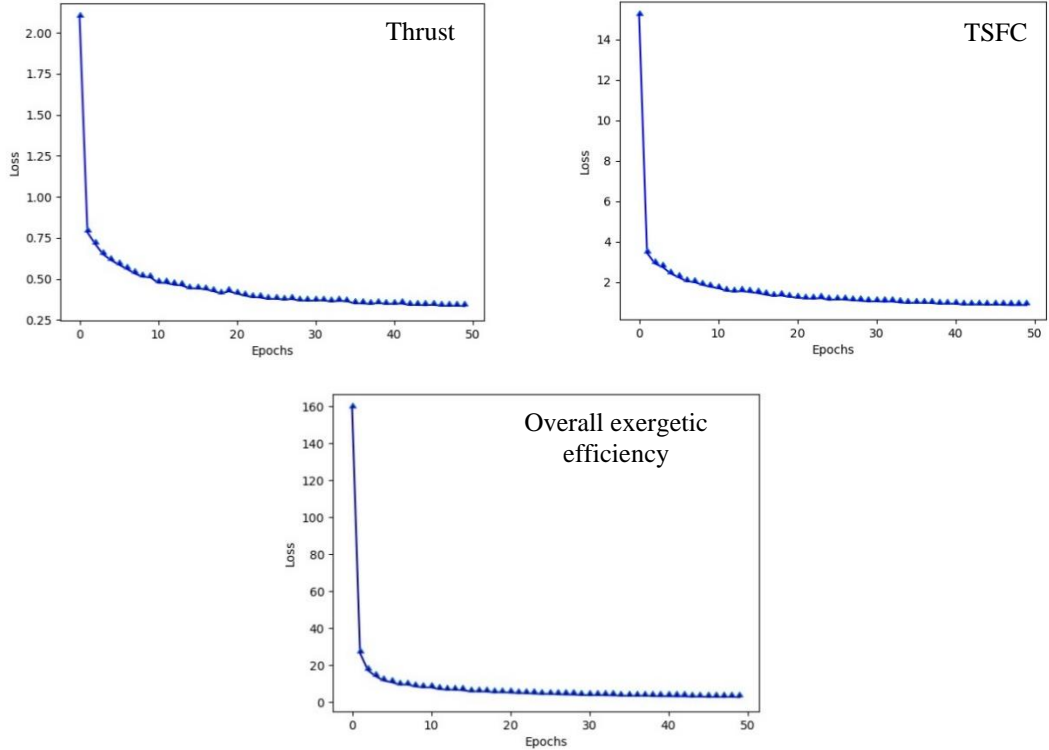

**Fig. 11.** Loss function variations with respect to Epochs for outputs variables

Figure 13 shows the error distribution of the 1520 test sample. For the thrust, 1347 samples (88.61%) are in the error range ± 1. Also, for the TSFC and exergetic efficiency, 1424 and 1381 samples, respectively, equivalent to 93.68% and 90.85% of the data are in the same error range. This high error distribution in the near-zero range indicates the success of the deep neural network in predicting test data.

Also, in order to measure the accuracy of the obtained model, the correlation factor ($R$), determination factor ($R^2$), root mean square error (RMSE), mean squared error (MSE), mean absolute error (MAE), and mean absolute percentage error (MAPE) are utilized. The values of RMSE, MSE, MAE express the difference between the predicted outputs and the true values. The closer these coefficients are to zero, the lower the error (higher accuracy) of the model.

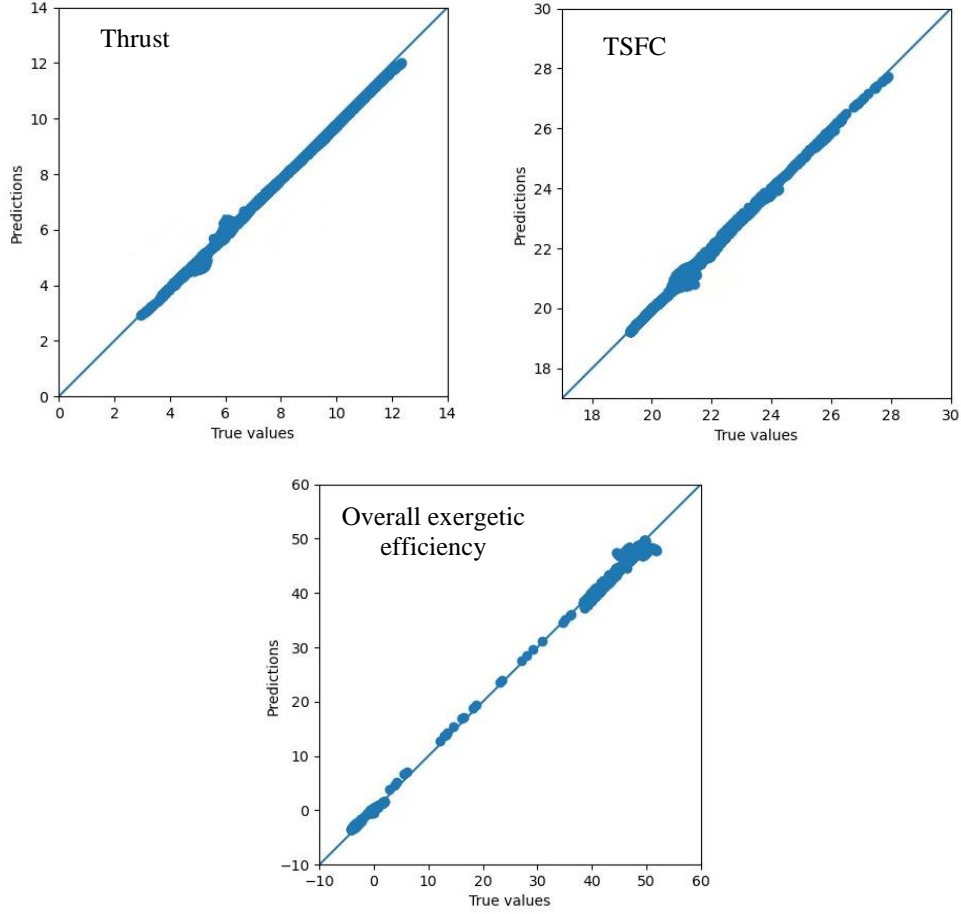

**Fig. 12.** predicted output values with respect to true values

The correlation and determination coefficients represent the correlation between the true values and the data predicted from the deep neural model. The closer the values of $R$ and $R^2$ to one, the closer the estimated values are to the true values. Finally, the Mean absolute percentage error provides a good indication for the accuracy of the obtained model. In general, the RMS, MSE, and MAE criteria depend on the amplitude of the output values, while the $R$, $R^2$, and MAPE coefficients can ignore the amplitude effect and provide a relative comparison. All the desired indicators are defined as follows:

$$R = \frac{\sum_{i=1}^{n}[(y_i - y_i^{mean})(\hat{y}_i - \hat{y}_i^{mean})]}{\sqrt{[\sum_{i=1}^{n}(y_i - y_i^{mean})^2][\sum_{i=1}^{n}(\hat{y}_i - \hat{y}_i^{mean})^2]}} \tag{58}$$

$$R^2 = \frac{\left[\sum_{i=1}^{n}(y_i - y_i^{mean})(\hat{y}_i - \hat{y}_i^{mean})\right]^2}{\left[\sum_{i=1}^{n}(y_i - y_i^{mean})\right]\left[\sum_{i=1}^{n}(\hat{y}_i - \hat{y}_i^{mean})\right]} \tag{59}$$

$$RMSE = \sqrt{\frac{1}{n}\sum_{i=1}^{n}(y_i - \hat{y}_i)^2} \tag{60}$$

$$RMSE = \frac{1}{n}\sum_{i=1}^{n}(y_i - \hat{y}_i)^2 \tag{61}$$

$$MAE = \frac{1}{n}\sum_{i=1}^{n}|y_i - \hat{y}_i| \tag{62}$$

$$MAPE = \frac{100\%}{n}\sum_{i=1}^{n}\left|\frac{y_i - \hat{y}_i}{y_i}\right| \tag{63}$$

where $y_i$ is the true values for i-sample, $\hat{y}_i$ is the output predicted by the deep model for i- sample, $y_i^{mean}$ is the mean of true values and $\hat{y}_i^{mean}$ is the mean of predicted data. The index is listed for two training and test sections as shown in table 4. To predict the first output, it is observed that the indicators $R$ and $R^2$ are 0.96 and 0.93, respectively. Also, the values of RMSE, MSE, and MAE are in the range of 0.5 and even lower. Also, the MAPE is about 5%; due to the mentioned values, the accuracy of the deep neural network is evaluated very high. For the second output, $R$ and $R^2$ are about 0.93 and 0.86, respectively. Error index values are also low and MAPE is about 1.5%. As a result, the success of the deep learning method in predicting the second output is also evident. Finally, for the third output, the $R$ and $R^2$ indices are 0.99 and 0.99, respectively, which is very close to 1, indicating a very high correlation between the predicted and the true values. Also, the MAPE is calculated below 3%. As a result, it can be seen that the highest correlation coefficients belong to the third output, and the lowest MAPE value belongs to the second one. The first output also achieves a balance between the two indices. in general, the network accuracy of all outputs is evaluated very high.

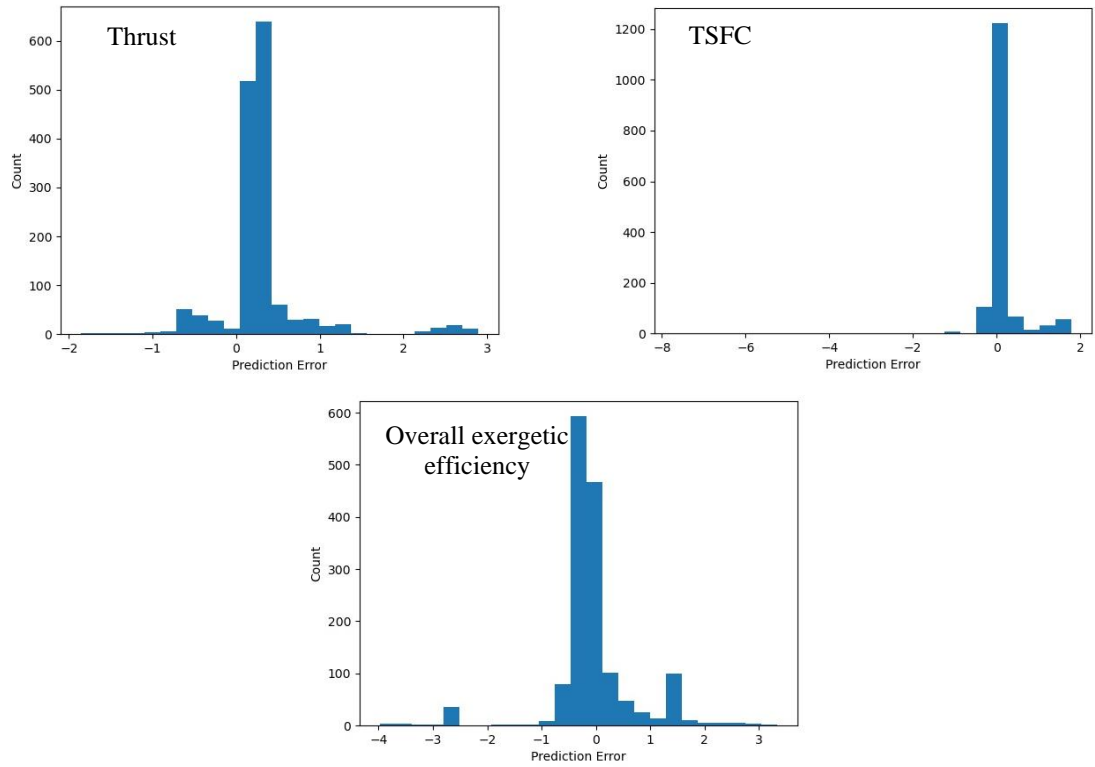

**Fig. 13.** Error distribution of the test samples for output variables

**Table 4.** Indicator values in the learning and testing sections

|  | **Thrust** | | **TSFC** | | **Overall exergetic efficiency** | |
|---|---|---|---|---|---|---|
|  | Train | Test | Train | Test | Train | Test |
| $R$ | 0.967579 | 0.968642 | 0.9302 | 0.927566 | 0.998204 | 0.998232 |
| $R^2$ | 0.936209 | 0.938267 | 0.865271 | 0.860378 | 0.996411 | 0.996467 |
| RMSE | 0.54496 | 0.536653 | 0.700278 | 0.696604 | 0.828118 | 0.811836 |
| MSE | 0.296981 | 0.287997 | 0.490389 | 0.485257 | 0.68578 | 0.659077 |
| MAE | 0.302363 | 0.301263 | 0.32562 | 0.316368 | 0.56414 | 0.560377 |
| MAPE | 5.121589 | 5.022951 | 1.471306 | 1.439301 | 2.115357 | 2.921235 |

## 5. Conclusion

In the present study, the thermodynamic analysis and modeling of the F135 PW100 engine have been performed in two different phases. In the first phase, the effect of several parameters on TSFC, thrust, intake air mass flow rate, thermal and propulsive efficiency, exergetic efficiency, and the exergy destruction rate is investigated. Based on the attained results, in the second phase, a deep neural model is obtained to predict thrust, TSFC, and overall exergetic efficiency.

The first phase observations show that, by reducing the intake air temperature 10 K at flight condition of 30,000 m altitude and Flight-Mach number of 2, Thrust and thermal efficiency are increased 6.8% and 1.14, respectively, also, TSFC and propulsive efficiency are decreased by 2.48% and 1.81%. In the checking fuel types, by using the hydrogen fuel compared to JP10 fuel at a flight altitude of 30,000 m and Flight-Mach number of 2.5, the thrust and thermal efficiency are increased 10.44% and 6.46% respectively, also the TSFC and propulsion efficiency are decreased in 64.84% and 2.16% respectively. The results of the second phase demonstrate that, the correlation factor for prediction of thrust, TSFC, and overall exergetic efficiency are calculated as 0.96, 0.93, and 0.99, respectively. Moreover, the mean absolute percentage error for the aforementioned outputs are 5.02%, 1.43%, and 2.92%. Accordingly, it is concluded that, the accuracy of attained deep neural model for the prediction of all outputs is very high.